\documentclass{article}

\usepackage{amsmath}
\usepackage{amssymb}
\usepackage{caption}
\usepackage{comment}

\usepackage[preprint]{neurips_2024}

\usepackage{placeins}

\usepackage[utf8]{inputenc} 
\usepackage[T1]{fontenc}    
\usepackage{hyperref}       
\usepackage{url}            
\usepackage{booktabs}       
\usepackage{amsfonts}       
\usepackage{nicefrac}       
\usepackage{microtype}      
\usepackage{xcolor}         
\usepackage{tcolorbox}
\usepackage{enumitem}
\usepackage{tikz}
\usetikzlibrary{shapes.geometric, arrows.meta, positioning}

\title{Tiny-TSM: Efficiently Training a Lightweight SOTA Time Series Foundation Model}

\author{%
Felix Birkel\\
Prior Labs \thanks{Work done as an independent researcher before joining Prior Labs.} \\
  \texttt{fpb23@cantab.ac.uk} \\
}

\begin{document}

\maketitle

\begin{abstract}
We present Tiny-TSM, a time series foundation model characterized by small scale, economical training, and state-of-the-art performance. It comprises $23$M total parameters, trained on a single A100 GPU in less than a week using a new synthetic data generation and data augmentation pipeline (SynthTS). Without any neural architecture search, hyperparameter tuning, or scaling up model size, Tiny-TSM achieves state-of-the-art performance on a wide range of time series benchmark datasets, often outperforming much larger models and even matching the performance of much larger, industrial-scale, likely highly tuned foundation models. Specifically, Tiny-TSM outperforms all other time series foundation models we evaluated on medium- and long-term forecasting tasks under MSE loss, while short-term accuracy is still competitive with state-of-the-art models.

We also introduce a causal input normalization scheme that enables time series models to be trained with dense next-token prediction loss, significantly accelerating convergence speed and reducing training time. 

All experiments were conducted on a single A100 GPU, illustrating the practicality of the proposed approach in a resource-constrained setting.

\end{abstract}


\begin{figure}[!htbp]
    \centering
    \includegraphics[width=0.9\linewidth]{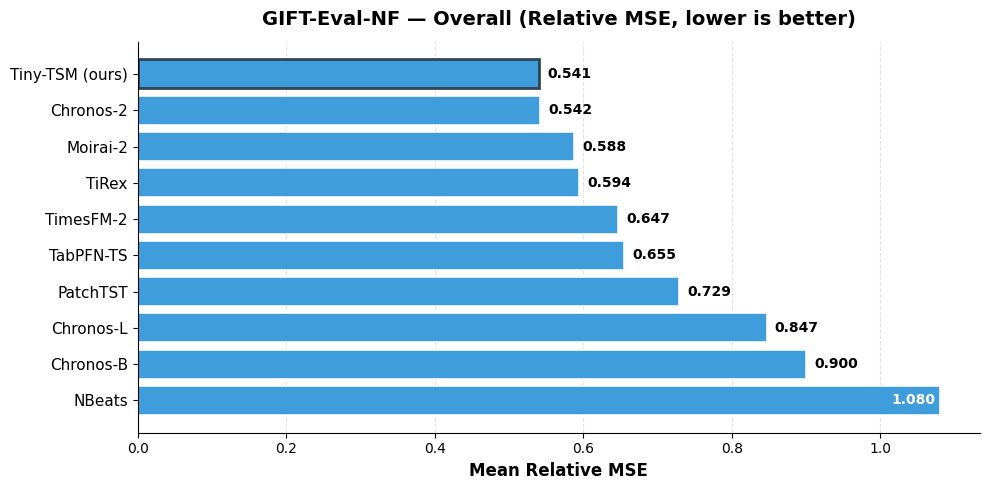}
    \caption{Tiny-TSM’s relative MSE performance on the GIFT-Eval-NF benchmark compared to other state-of-the-art time series foundation models.}
    \label{fig:bench1}
\end{figure}

\section{Introduction}

Time series analysis is a key domain for modern data science, with applications in healthcare \citep{Che2016RecurrentNN}, economics \citep{StockWatson2002}, climate and environmental sciences \citep{dimri2020time}, energy and power systems \citep{gasparin2022deep, nabavi2024deep}, transportation and mobility \citep{liu2020traffic}, industrial IoT and monitoring \citep{liu2020deep, belay2023unsupervised}, retail and supply chain management \citep{Spyros2021, sartipi2025benchmarking}, and astronomy \citep{kugler2015featureless, cabrera2024atat}. 
A modern deep-learning-based approach for time series prediction has attracted growing interest, with an increasing number of publications on time series foundation models \citep{ansari2024chronoslearninglanguagetime, hoo2025tablestimetabpfnv2outperforms, das2024decoderonlyfoundationmodeltimeseries, cohen2024tototimeseriesoptimized, auer2025tirexzeroshotforecastinglong}. Existing work on time series foundation models follows a variety of training paradigms, including masked modeling, fixed-horizon prediction, and other task-specific objectives.

In this work, we investigate how far one can push the performance of a \emph{small}, compute-efficient time-series foundation model trained primarily on synthetic data. We leverage dense next-token prediction together with multi-target forecasting to maximize training signal, and introduce a causal normalization scheme that enables such dense objectives without test leakage. Our main result is that \emph{Tiny-TSM}, a 23M-parameter encoder-based model trained using our \emph{SynthTS} framework, achieves state-of-the-art performance on a broad suite of benchmark datasets, rivaling much larger models trained on curated sets of real world data.

This technical report focuses on the design choices that make this possible, rather than exhaustive neural architecture search or large-scale hyperparameter tuning: All work was carried out as a one-person side project on a single GPU under tight time and budget constraints.

\subsection{Contributions}

\begin{itemize}
  \item We introduce \textbf{DART-Norm} (Drift Aware Rolling Timeseries Normalization, section~\ref{sec:dart}), a causal input normalization scheme for time series that enables dense next-token prediction losses without test leakage and exposes normalization drift explicitly to the model.
  
  \item We propose \textbf{SynthTS} (section~\ref{sec:synthetic-data}), a pipeline that generates realistic synthetic time series data and allows for seamless augmentation of real-world time series.
  
  \item We showcase the effectiveness of the above proposed methods in \textbf{Tiny-TSM}, a small, patched encoder-based foundation model for time series. Tiny-TSM matches or exceeds the performance of substantially more expensive and heavily tuned forecasters, achieving state-of-the-art performance across a broad range of benchmarks.


\item We introduce \textbf{Stride-Interleaved Forecast Inference} (SIFI, section~\ref{sec:inference}), an inference-time ensemble technique that enables models trained with a fixed maximum context length to exploit longer historical series by forecasting on downsampled views and interleaving their predictions. We also demonstrate the benefit of multivariate feature augmentation, where derived transforms of the target variable are added as additional channels to boost accuracy, especially for originally univariate series.

\item We showcase the effectiveness of using a \textbf{coarse-grid loss} (section~\ref{sec:train_det}) to enable strong medium- and long-term forecasting performance. Indeed, Tiny-TSM outperforms all other foundation models we tested in medium- and long-term forecasting tasks, largely due to coarse-grid loss and SIFI.
    
\end{itemize}

\begin{figure}
    \centering
    \includegraphics[width=
    1.0\linewidth]{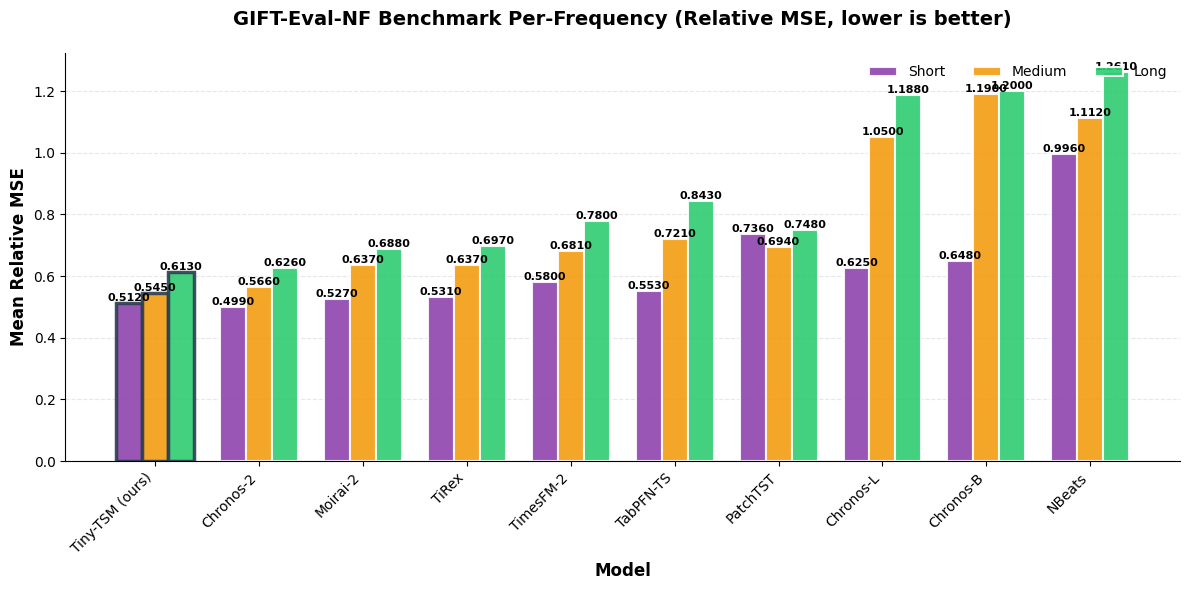}
    \caption{Per-prediction-length results on GIFT-EVAL-NF}
    \label{fig:BenchPerHzn}
\end{figure}

\section{Background}
\label{gen_inst}

\subsection{Time series forecasting}

Let $(x_t)_{t \ge 0}$ denote a univariate time series, and let $x_{a:b}$ denote the subsequence $(x_a, x_{a+1}, \dots, x_b)$. The time-series forecasting problem is to predict future values given past observations, typically by approximating the conditional distribution
\begin{equation}
    p(x_{t+1:T} \mid x_{0:t}).
\end{equation}

In many practical applications, a \emph{point forecast} suffices, in which case the goal is to estimate the conditional mean
\begin{equation}
    \mathbb{E} \big[ x_{t+1:T}\  | \ x_{0:t} \big]
    \label{eq:1}
\end{equation}
in case of a mean squared error (MSE) loss, or the corresponding conditional median for a mean absolute error (MAE) loss.

Time series may be univariate ($x_t \in 	\mathbb{R} \  \forall t$) or multivariate ($x_t \in 	\mathbb{R}^n \  \forall t$). In multivariate time series forecasting, one often aims to predict a single \textit{target} covariate - the remaining covariates can be useful to inform predictions for the target, but need not be predicted themselves. Let $x_{m,t}$ denote the target variable and $x_{1:n,t}$ the full covariate vector at time $t$. The corresponding conditional mean in Eq.~\eqref{eq:1} becomes 
\begin{equation}
    \mathbb{E}\big[x_{m,t+1:T} \mid x_{1:n,0:t}\big].
\end{equation}

More generally, some covariates may be known into the future (e.g., calendar features). Let $I \subset \{1,\dots,n\}$ index the covariates with known future values. The forecasting problem is then
\begin{equation}
    \mathbb{E} \big[ x_{m, t+1:T} \ | \  x_{1:n, 0:t}\ , \  x_{I,t+1:T} \big], 
    \label{eq:3}
\end{equation}
This setting includes the previous one as a special case (no future known covariates), hence we focus on this setting for this report. 

\begin{figure}
    \centering
    \includegraphics[width=1.0\linewidth]{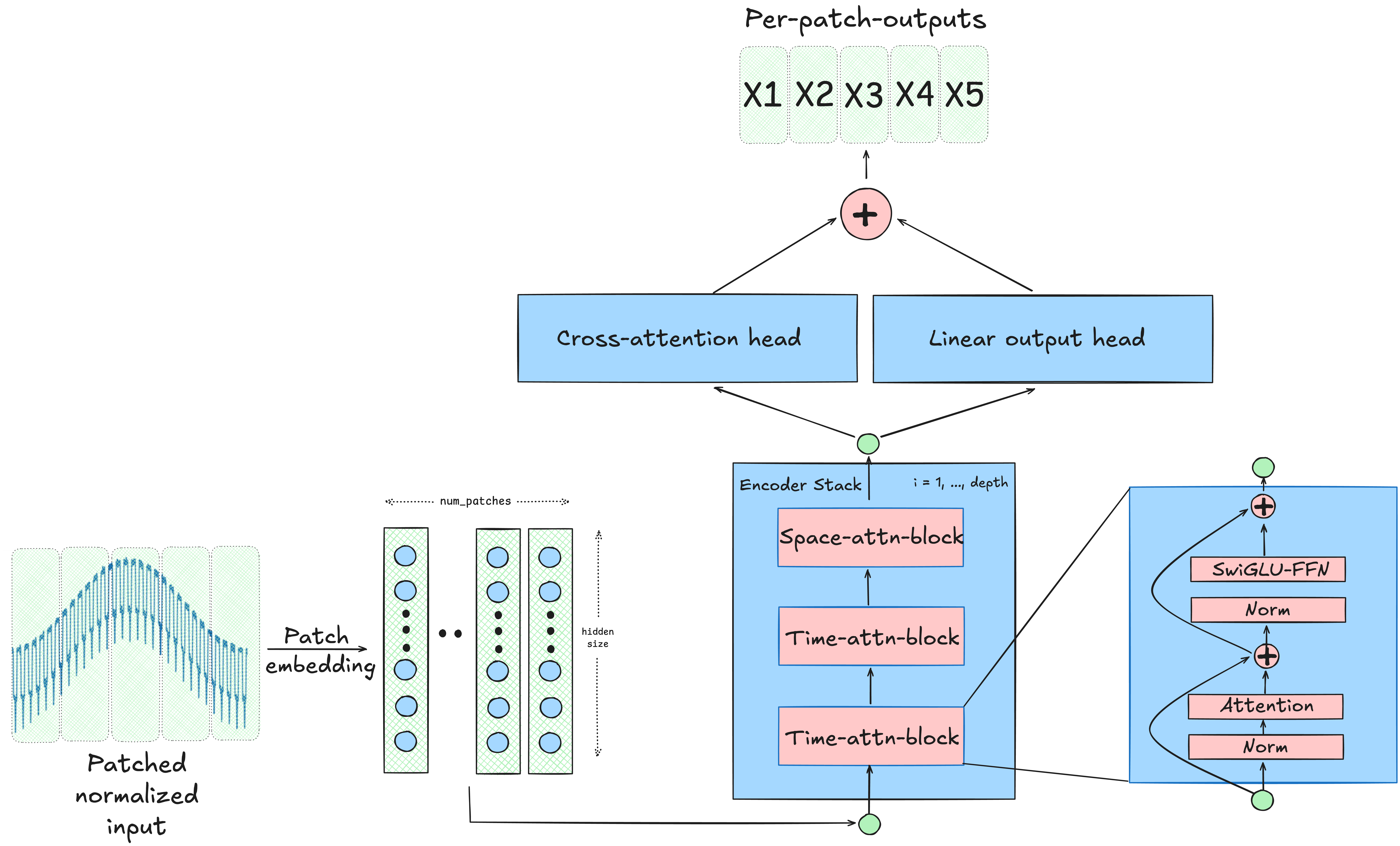}
    \caption{The Tiny-TSM architecture: We employ a patched encoder-based architecture. First, a linear patch embedding projects patches into the model's hidden dimension. In the encoder stack, we interleave time- and feature-attention blocks at a 2:1 ratio. The internal structure of an attention block is shown on the right. The output of the encoder stack is processed by two forecast heads: A cross-attention head and a linear head, the outputs of which are added together to form the final per-patch predictions.}
    \label{fig:Arch}
\end{figure}

\subsection{Time series foundation models}

\begin{figure}
    \centering
    \includegraphics[width=1.0\linewidth]{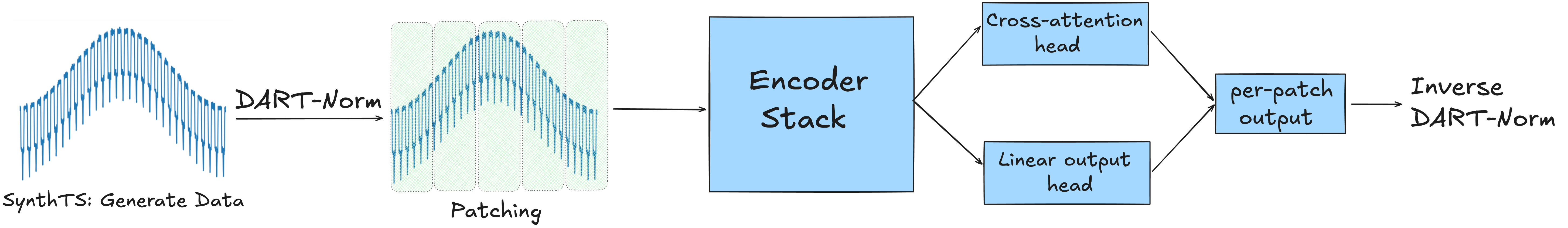}
    \caption{The Tiny-TSM pipeline. Data is first normalized using DART-Norm, then patched and consumed by the model. The model produces per-patch predictions, all of which are then de-normalized.}
    \label{fig:Pipeline}
\end{figure}

The foundation model paradigm has received strong attention in recent years, originally popularized by large language models (LLMs) in NLP~\citep{bommasani2022opportunitiesrisksfoundationmodels}.
Foundation models are powerful pre-trained models that solve downstream tasks via in-context learning~\citep{brown2020languagemodelsfewshotlearners} and/or light fine-tuning. 

The foundation model paradigm has since been extended beyond NLP to computer vision~\citep{he2021maskedautoencodersscalablevision, radford2021learningtransferablevisualmodels}, tabular data~\citep{hollmann2023tabpfntransformersolvessmall, grinsztajn2025tabpfn25advancingstateart}, and time series modeling~\citep{das2024decoderonlyfoundationmodeltimeseries,ansari2024chronoslearninglanguagetime}. 
Models such as TabPFN~\citep{hollmann2023tabpfntransformersolvessmall} showcase the potential of foundation models for tabular and time series data, in many cases outperforming specialized models such as gradient boosting machines.

In recent years, many time series foundation models have been proposed, typically relying on either fixed-length forecasting, masked modeling or autoregressive modeling objectives. 
Other approaches adapt pre-trained LLMs to model time series directly, for example by tokenizing continuous sequences and leveraging existing language backbones for forecasting and representation learning.

Despite rapid progress, many current time-series foundation models are large and expensive to train, and their performance often depends on curated real-world datasets and substantial computational budgets. This motivates the question addressed in this work: \emph{How competitive can a small, synthetic-data-driven foundation model be if equipped with the right normalization, data generation, and inference strategies?}

\section{Related Work}

\paragraph{Deep learning methods for time series.}
Early deep learning methods for time series forecasting focused on sequence models such as recurrent networks and temporal convolutional networks. More recently, purely feedforward architectures such as N-BEATS~\citep{oreshkin2020nbeatsneuralbasisexpansion} have shown that carefully designed multilayer perceptrons with backward/forward residual links can attain state-of-the-art accuracy on classical benchmarks. Transformer-based models have since become dominant for long-horizon forecasting. PatchTST~\citep{nie2023timeseriesworth64} introduced a patched time series Transformer, in which each univariate series is segmented into overlapping patches that serve as tokens, reducing attention complexity. Today, patching is commonplace in transformer-based time series models. 

\paragraph{Time series foundation models.} The idea of large, pre-trained foundation models applied in a zero-shot manner has recently become popular for time series. Chronos~\citep{ansari2024chronoslearninglanguagetime} tokenizes continuous values via scaling and quantization and trains T5-style language models autoregressively on time-series “text”. TimesFM~\citep{das2024decoderonlyfoundationmodeltimeseries} adopts a decoder-only Transformer that operates directly on continuous inputs. Moirai~\citep{woo2024unifiedtraininguniversaltime} proposes a masked encoder-based Transformer trained on the LOTSA corpus, with frequency-specific projection layers and any-variate attention. Parallel to these, tabular foundation models such as TabPFN have been successfully adapted to time series forecasting (TabPFN-TS) by reframing forecasting as a tabular regression problem and using light feature engineering~\citep{hollmann2023tabpfntransformersolvessmall,hoo2025tablestimetabpfnv2outperforms}. 

\paragraph{Synthetic time series data and data augmentation.} Synthetic data and data augmentation have emerged as key ingredients for training robust time series and tabular foundation models. TabPFN is trained entirely on synthetic tabular tasks and nevertheless transfers effectively to time series when combined with appropriate features~\citep{hollmann2023tabpfntransformersolvessmall,grinsztajn2025tabpfn25advancingstateart}. CauKer~\citep{xie2025caukerclassificationtimeseries} proposes a synthetic data pipeline based on Gaussian process kernel composition and structural causal models to pretrain classification TSFMs purely on synthetic sequences. Chronos-2 likewise leverages mixtures of real and synthetic datasets in its training.

\section{Synthetic Data}
\label{sec:synthetic-data}

\begin{figure}
    \centering
    \includegraphics[width=1.0\linewidth]{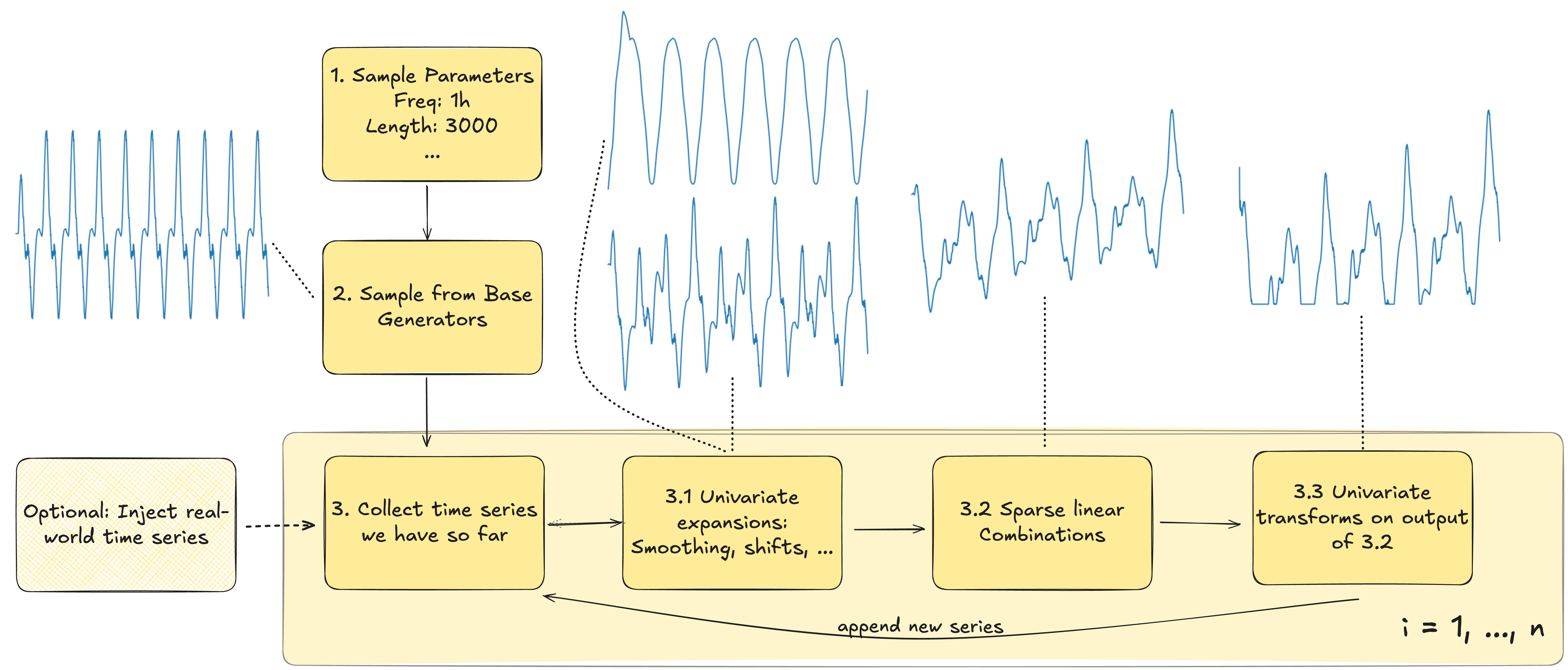}
    \caption{Overview of the SynthTS synthetic data generation and augmentation framework. We visualize a simplified example of a single single series sampled from a base generator and how it is modified by univariate expansions, sparse feature mixing, and post-transforms.}
    \label{fig:synthts}
\end{figure}

Recent work has shown the vast potential of training on carefully designed synthetic data, as demonstrated by the TabPFN family of models ~\citep{hollmann2023tabpfntransformersolvessmall, grinsztajn2025tabpfn25advancingstateart}. For training Tiny-TSM, we introduce \emph{SynthTS}, a synthetic time series generation and augmentation framework designed to mimic key properties of real-world series while covering a broad range of temporal patterns.
Figure~\ref{fig:synthts} illustrates the overall data generation procedure.

At a high level, SynthTS first samples univariate time series from a family of base series generators based on a randomly chosen time-index and then combines these primitive series through a sequence of composition and transformation steps. The same augmentation pipeline can be applied to a mix of sampled synthetic and real-world time series, allowing synthetic and real data to be mixed seamlessly.

\paragraph{SynthTS.}

\paragraph{1. Frequency and parameter sampling.}
Each synthetic series is associated with a time index and a base observation frequency (e.g., hourly). We sample a base frequency and a random start time to construct the time index. We then sample a set of \emph{compatible} frequencies that are integer multiples of the base frequency, assigning higher weights to naturally occurring scales. For example, for an hourly base frequency we assign higher probability to daily (24h) and weekly (168h) multiples, while still allowing other integer multiples. We also allow the time index to be absent, to simulate pure sequence data.

In addition, we sample other batch-level parameters such as sequence length and noise level.

\paragraph{2. Sampling from base generators.}
We define a collection ( $\sim 50$) of \emph{base generators}, each of which samples from a specific class of time series given a set of input parameters. 
Examples include a sinusoidal generator (with random phase and amplitude, given a frequency) or a random polynomial generator. 
We implement a mixture of periodic, polynomial, and noise-based base generators, along with several additional variants. 
A comprehensive overview with visualizations is provided in the appendix.

\paragraph{3. Generating derived time series.}
Starting from the base series sampled in step~2, we apply a multi-step augmentation pipeline. Real-world time series can be incorporated simply by appending them to the base pool, provided lengths and frequencies are matched.

We repeat the following steps for a chosen number of augmentation rounds:

\subparagraph{3.1 Univariate expansions.}
For each input time series, we compute a randomly chosen number $N$ ($N \in \{1, \dots, 5\}$) of univariate transforms. These transforms operate along the time axis and include shifts, kernel smoothing, and autoregressive transforms.

\subparagraph{3.2 Sparse linear combinations along the feature-dimension.}
We sample random weights to form a set of sparse linear combinations of the expansions from step~3.1.
Sparsity is crucial: dense mixtures tend to blur patterns and reduce diversity. Optional noise can be added at this stage.

\subparagraph{3.3 Post-transforms.}
Finally, we apply one of many \emph{post-transforms} focusing on pointwise operations and support effects. Examples include
\[
x \mapsto \mathrm{ReLU}(\mathrm{MinMaxScaler}(x) - 0.5),
\]
to mimic series with a natural floor at zero, or 
\[
x_1 \mapsto x_1 \cdot \mathrm{MinMaxScaler}(x_2),
\]
to model amplitude modulation. 
Other post-transforms inject missing values, outliers, or random spikes.

The outputs of step~3.3 are appended to our set of input time series, the intermediate outputs from steps~3.1 and~3.2 are discarded. 

In practice, we use between 2 and 5 augmentation rounds and subsample a fixed number of time series for each round from the current pool of inputs, in order to avoid an exponential growth in the number of derived series. 

\section{Input normalization and enabling next-token loss}
\label{sec:dart}

Time-series and tabular foundation models must handle inputs with widely varying scales. A common solution is to standardize each series (e.g., subtracting its mean and dividing by its standard deviation) before feeding it to the model, and to apply the inverse transform to the model output. In the time-series setting, however, such normalization interacts in important ways with non-stationarity and with dense next-token prediction objectives.

\begin{figure}
    \centering
    \includegraphics[width=1.0\linewidth]{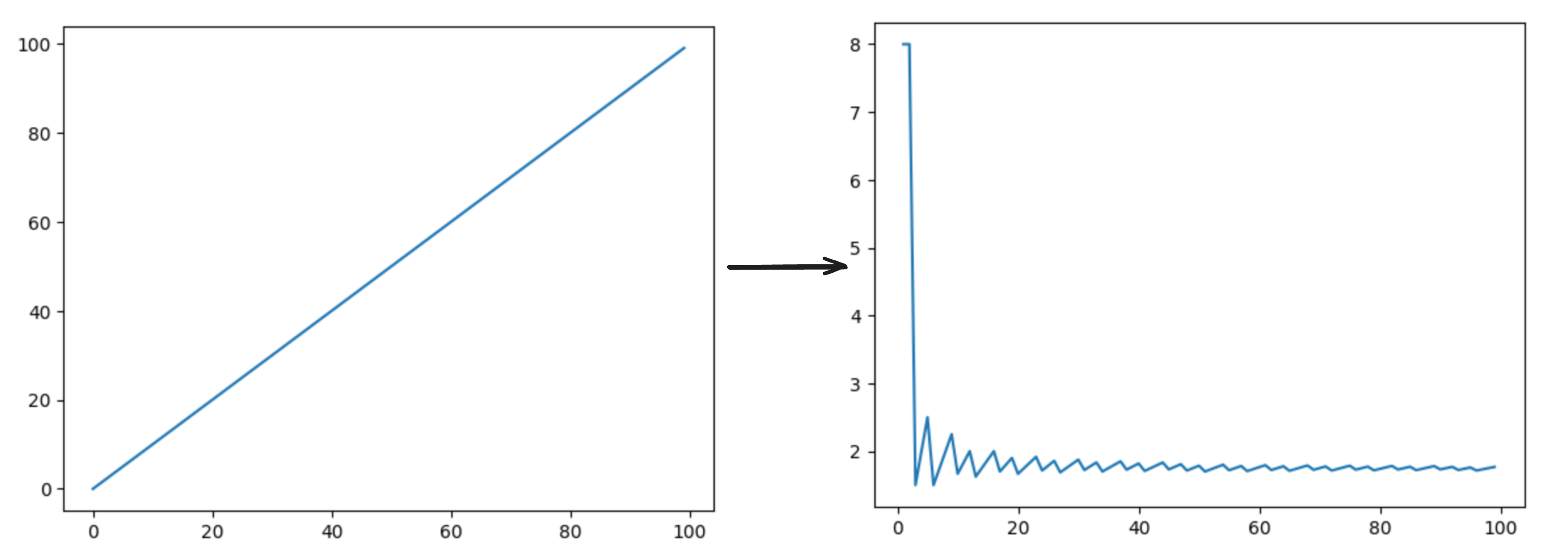}
    \caption{Rolling standard scaling turns a linear trend (shown on the left) into an eventually constant series (shown on the right), destroying identifiability. Therefore, providing the model with normalization drift metrics is essential when using rolling standard scaling.}
    \label{fig:nonident}
\end{figure}

First, many time series datasets exhibit strong non-stationarity, so scaling parameters estimated on the training portion of the series (often the early part) may not generalize well to future points. 
This is particularly evident in the presence of exponential or otherwise super-linear trends. Rolling standardization can address this, but can also lead to non-identifiability as illustrated in Figure \ref{fig:nonident}. 

Second, per-series standard scaling complicates dense next-token training. If normalization statistics are computed using both context and target points, then using those targets in the loss leaks information. Restricting normalization to a fixed prefix (e.g., the first 80\% of each series) avoids leakage but sacrifices the dense training signal provided by next-token objectives.

To address these issues, we introduce \emph{DART-Norm}, a causal normalization scheme that enables dense next-token prediction while (i) ensuring strictly causal normalization with no leakage, and (ii) mitigating artifacts of rolling normalization by exposing stepwise changes in the normalization statistics to the model.

\begin{figure}
    \centering
    \includegraphics[width=1.0\linewidth]{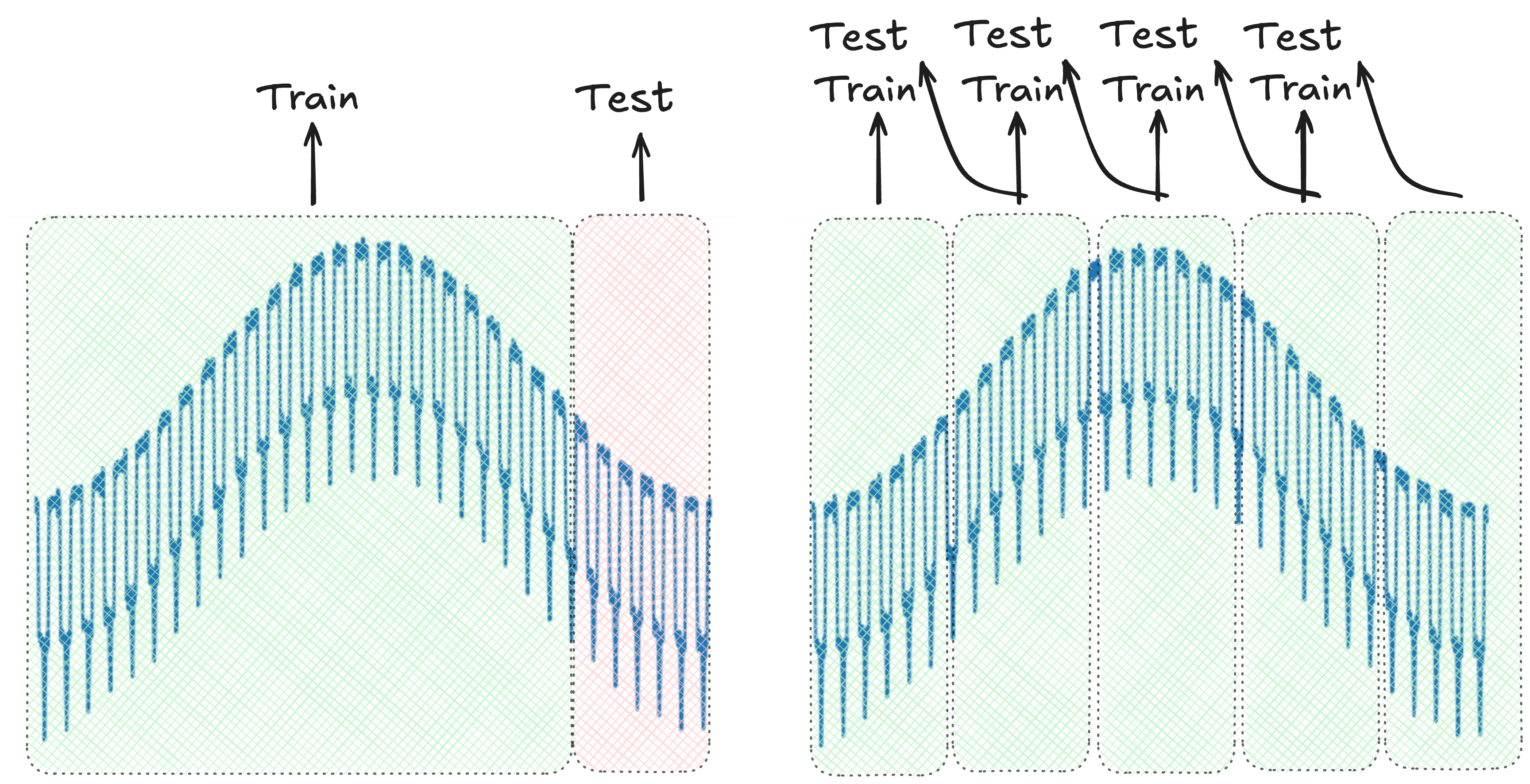}
    \caption{Comparing a test-at-end setup with a next-token-prediction setup. In the test-at-end setup, the train data can be used to normalize the input sequence. This is not directly possible in the next-token-prediction setup without causing test leakage or the need to do multiple forward passes.}
    \label{fig:nextpatch}
\end{figure}

\paragraph{Algorithm: DART-Norm}

\paragraph{1. Rolling standardization.}
Given a time series $(y_t)$, we first compute its rolling mean and rolling standard deviation:
\begin{equation}
    m_t = \mathrm{mean}(y_{0:t})
\end{equation}
\begin{equation}
    s_t = \mathrm{std}(y_{0:t})
\end{equation}
and define the rolling-standardized series
\begin{equation}
    x_t = \frac{y_t - m_t}{s_t}.
\end{equation}

\paragraph{2. Drift features.}
Next, we compute two drift metrics that track changes in the normalization statistics over time: a scaled stepwise difference in the mean, and a logarithmic relative stepwise difference in the standard deviation:
\begin{equation}
    d_t = \frac{m_t - m_{t-1}}{s_{t-1}}
\end{equation}
\begin{equation}
    r_t = \log\!\left(\frac{s_t}{s_{t-1}}\right) = \log s_t - \log s_{t-1}.
\end{equation}
We then concatenate $x_t$, $d_t$, and $r_t$ to form the model inputs at each time step.

\paragraph{3. Next-token targets.}
Finally, we construct next-token prediction targets \emph{anchored} to the most recent in-context normalization. 
For a given context length $T$ and prediction horizon $h$, the targets are
\begin{equation}
    \frac{y_{T+1:T+h} - m_T}{s_T}.
\end{equation}

Empirically, DART-Norm enables dense next-token training with strictly causal normalization and we observe a roughly \textbf{3$\times$} speed-up in convergence, as well as stronger final model performance, compared to a masked ``test-at-the-end'' training setup. 
We note however that under a sufficiently large computational budget, we would not necessarily expect a model trained with DART-Norm to converge to a better checkpoint than an otherwise identical model trained with a masked test-at-the-end objective, given that there are no benefits to this normalization scheme in fixed-input-length forecasting tasks that are common in benchmarks - the benefits are in enabling dense gradient signals during training.

Simply feeding $x_t$ to the model is not sufficient, since at inference the model must produce forecasts in the original scale of $y_t$, not $x_t$ (see Figure \ref{fig:nonident}). 
Unlike raw input time series, the drift features $d_t$ and $r_t$ naturally exhibit comparable scales across different series and require no further normalization beyond optional clipping. Moreover, given $(x_t, d_t, r_t)$, one can reconstruct $y_t$ up to a global affine transform, implying that the model can internally emulate fixed-window normalizations if beneficial.

In online settings where new data points arrive continuously, DART-Norm also makes it easy to incorporate fresh observations into the training context with minimal overhead, allowing for inexpensive updates to the key-value-cache.

\section{Model Architecture}
\label{sec:arch}

We adopt a standard encoder-based patched time series transformer, following the architecture introduced in Toto~\citep{cohen2024tototimeseriesoptimized}. 
We use a patch length of 32 to keep computation fast; due to time constraints, we did not explore alternative patch sizes. 
On top of the base architecture, we add a few small adaptations:
\begin{itemize}
    \item We add learnable embeddings for missing values of shape (patch length, hidden size).

    \item We add a lightweight cross-attention head that attends to known future covariates, in addition to a linear head, producing patch-level forecasts.

    \item We remove the initial normalization layer (RMSNorm) prior to the first encoder layer, as normalizing all input patches to a constant scale removes valuable patch-level information.

    \item We prepend a small number of learnable pad tokens to handle short sequences more effectively.

    \item We use a 2:1 ratio of time-mixing attention layers to space-mixing attention layers, reflecting the stronger temporal coupling in typical forecasting tasks.
    
\end{itemize}

A general overview of the model architecture is described below and visualized in figure~\ref{fig:Arch}: 

\begin{enumerate}
    \item Patching: For a given patch length (32 in our case), we chunk the inputs into patches of length $patch\_len$: 
    \begin{equation} 
    (x_t)_{t = 0, ..., T} \in \mathbb{R} \  \mapsto \  (\bar{x}_{i,j}) \ \  where \ \ \bar{x}_{i,j} = x_{i*32 + j} \ \ \ for  \ j = 1,... , 32,
    \end{equation}
    padding or discarding early time steps as needed. Patches for time series models are analogous to tokens in NLP. 

    For multivariate series, we patch (and embed) all channels independently.
    
    \item Patch embedding: We embed each patch into the model's hidden dimension using a linear projection. We concatenate the drift features produced by DART-Norm with the input as described earlier, thus our patch embedding projects $3 \times patch\_length \mapsto hidden\_size$.

    \item Encoder stack: We iteratively apply standard post-norm transformer layers, interleaving multi head attention (MHA), residual connections, feedforward networks (FFN) and normalization layers (RMSNorm).

    \begin{equation}
        RMSNorm \mapsto MHA \mapsto Residual \mapsto RMSNorm \mapsto FFN \mapsto Residual
    \end{equation}
    
    For maximum expressivity, we use SwiGLU activations in the FFNs, where 
    \begin{equation}
    \begin{split}
        & SwiGLU(x) = Swish_{W_3,\beta}(W_1 x + b) \otimes (W_2 x + c)\\
        & Swish_{W,\beta}(x) = W(x\sigma(\beta x))
    \end{split}
    \end{equation}
    
    for learnable matrices $W_{1,2,3}$ and parameters $b,c,\beta$. We apply one spatial transformer layer after every two temporal transformer layers. Temporal attention layers are causal, while spatial ones are not.

    \item Output layer. To leverage known future covariates (including time-index-derived features) the patch-level encoder hidden representations feed into two lightweight forecast heads whose outputs are summed:
    \subitem (i) a linear head mapping each patch representation directly to the forecast horizon, and
    \subitem (ii) a small covariate cross-attention head that refines predictions using the known future covariates.

\end{enumerate}

For multivariate time series, we predict all input variables jointly at each time step, which yields lower-variance gradients than univariate forecasting and naturally supports multi target prediction. 
Known future covariates are not forecast.

We use a point-forecasting setup with a simple Huber loss. One could instead implement a probabilistic forecast head (mixture-of-t-distribution head \citep{cohen2024tototimeseriesoptimized}; quantile forecast head \citep{auer2025tirexzeroshotforecastinglong}) but we did not explore this direction.

\section{Inference}
\label{sec:inference}

Enabled by the model's highly efficient architecture and the resulting very fast run-time, we can use a number of inexpensive test-time enhancements. We exploit both symmetry and the sequential nature of time series to reduce variance and improve accuracy.

\paragraph{Symmetry-based ensembling.}
We ensemble the model with a mirrored version of the input:
\begin{equation}
    \mathrm{output} = \frac{f(x_t) - f(-x_t)}{2},
\end{equation}
where $f$ denotes the model. This idea reduces variance of predictions by exploiting the internal non-linearity of the model that contrasts the linear input transform. Additional ensembles can be formed by adding small amounts of noise to the input (e.g., Gaussian noise with standard deviation equal to 1\% of the input's standard deviation).

\paragraph{Multivariate feature augmentation.}
Because Tiny-TSM is trained natively on multivariate inputs, we append transformed versions of the target variable as additional channels at inference time. Examples include signed-square and signed-square-root transforms, as well as kernel-smoothed variants. While the exact choice of transforms has limited impact, we find that adding several highly correlated channels consistently improves performance, likely by effectively increasing the model's representational capacity for the target.

\paragraph{Stride-Interleaved Forecast Inference.}
Tiny-TSM is trained with a maximum context length of 4096. To exploit longer historical series without changing the model, we introduce a downsample-and-interleave ensemble. 

Intuitively, SIFI runs the model on strided, downsampled views of the history (e.g. even timestamps vs odd timestamps) and then interleaves their predictions back onto the original time grid.

Given a series $(x_t)_{t=1}^T$ and a stride $n$, we construct $n$ downsampled series by taking every $n$-th point with different offsets:
\[
x^{(k)} = (x_{k}, x_{k+n}, x_{k+2n}, \dots), \qquad k = 1,\dots,n.
\]
We run the model on each downsampled series to obtain forecasts $\hat{x}^{(k)}$ at the corresponding coarse time indices, and then interleave these forecasts back onto the original time grid:
\[
\hat{x}_t = \hat{x}^{(k)}_{j}
\quad \text{for} \quad
t = k + jn,
\]

This method works well for time series with strong seasonality as well as smooth time series, but less well for time series with strong local dependencies and significant non-smooth local patterns, where skipping time-steps omits substantial information. Empirically, we observe significant reductions in loss on long-context series when applying SIFI.

A schematic is:

\begin{equation}
\begin{array}{rrrrrr}
   & \{x_t\}_{t=2,4,6,...} \mapsto & Model \mapsto & \{\hat{x}_t\}_{t=2,4,6,...} \mapsto & \searrow  &  \\
  \{x_t\}_{t=1, 2, 3, ...} & & & & & \{\hat{x}_t\}_{t=1,2,3,...}
  \\
   & \{x_t\}_{t=1,3,5,...} \mapsto & Model \mapsto & \{\hat{x}_t\}_{t=1,3,5,...} \mapsto & \nearrow &
\end{array}
\end{equation}

\section{Evaluation and Benchmarking}
\label{sec:eval}

We evaluate Tiny-TSM on all non-financial datasets from the GIFT-Eval benchmark \citep{aksu2024giftevalbenchmarkgeneraltime}, a collection of datasets we refer to as GIFT-Eval-NF. We compare against the leading time-series foundation models and staistical models as of the time of writing, using their per-dataset results from the GIFT-Eval git repository: Chronos (Base / Large) \citep{ansari2024chronoslearninglanguagetime}, Chronos-2 \citep{ansari2025chronos2univariateuniversalforecasting}, Moirai-2\citep{liu2025moirai20timeseries}, TiRex \citep{auer2025tirexzeroshotforecastinglong}, TimesFM-2 \citep{das2024decoderonlyfoundationmodeltimeseries}, TabPFN-TS \citep{hoo2025tablestimetabpfnv2outperforms}, PatchTST \citep{nie2023timeseriesworth64}, N-Beats\citep{oreshkin2020nbeatsneuralbasisexpansion}. For each dataset and model, we compute:
\begin{enumerate}
    \item Mean Relative MSE 
    \item Mean Relative MAE
\end{enumerate}
where ``relative'' denotes the ratio between the model's error and that of a seasonal naive baseline. Being a point-forecaster trained using a Huber loss, Tiny-TSM is adapted well to forecasting under squared error but less well for forecasting under other criteria. Results are shown in figures \ref{fig:bench1} and \ref{fig:BenchPerHzn}, as well as table \ref{tab:joint_mae_mse}.

\begin{table}[t]
\centering
\small
\setlength{\tabcolsep}{5pt}
\caption{Relative MSE and MAE per model and horizon.}
\label{tab:joint_mae_mse}
\begin{tabular}{lcccccccc}
\toprule
& \multicolumn{4}{c}{Relative MSE} & \multicolumn{4}{c}{Relative MAE} \\
\cmidrule(lr){2-5} \cmidrule(lr){6-9}
Model & Overall & Short & Medium & Long & Overall & Short & Medium & Long \\
\midrule
Tiny-TSM (ours) & \textbf{0.541} & 0.512 & \textbf{0.545} & \textbf{0.613} & 0.731 & 0.691 & 0.756 & 0.802 \\
Chronos-2 & 0.542 & \textbf{0.499} & 0.566 & 0.626 & \textbf{0.697} & \textbf{0.656} & \textbf{0.728} & \textbf{0.767} \\
Moirai-2 & 0.588 & 0.527 & 0.637 & 0.688 & 0.724 & 0.673 & 0.770 & 0.803 \\
TiRex & 0.594 & 0.531 & 0.637 & 0.697 & 0.723 & 0.674 & 0.764 & 0.802 \\
TimesFM-2 & 0.647 & 0.580 & 0.681 & 0.780 & 0.780 & 0.724 & 0.819 & 0.899 \\
TabPFN-TS & 0.655 & 0.553 & 0.721 & 0.843 & 0.767 & 0.703 & 0.813 & 0.879 \\
PatchTST & 0.729 & 0.736 & 0.694 & 0.748 & 0.856 & 0.842 & 0.856 & 0.890 \\
Chronos-L & 0.847 & 0.625 & 1.050 & 1.188 & 0.847 & 0.739 & 1.033 & 1.066 \\
Chronos-B & 0.900 & 0.648 & 1.190 & 1.200 & 0.890 & 0.750 & 1.064 & 1.079 \\
NBeats & 1.080 & 0.996 & 1.112 & 1.261 & 0.919 & 0.984 & 0.857 & 0.817 \\
\bottomrule
\end{tabular}
\end{table}

We did not evaluate any models on any datasets or benchmarks other than GIFT-Eval-NF at any point during this project.

\section{Training Details}
\label{sec:train_det}

Tiny-TSM is trained with a patch length of 32; a maximum input sequence length of 4096; a batch size of 4; a learning rate of $1 \times 10^{-4}$ (AdamW); and a maximum forecast horizon of 960 time steps.

To encourage strong performance across both short and long horizons, we subsample a maximum forecast horizon from $\{1,\dots,960\}$ for each training batch. On a subset of batches, we further apply a \textbf{coarse-grid loss}, computing the loss only on every $n$-th predicted time step for $n \in \{1, 2, 4, 8, 16, ..., 128\}$. This balances contributions from slow and fast dynamics and mitigates the dominance of noisy and high-variation series in the loss: A low-frequency sinusoid produces much smaller local errors than a high-frequency sinusoid under standard dense losses, leading to weaker gradients for the former.

All experiments were conducted on a single A100 GPU.

\section{Limitations and Extensions}

This work is intentionally limited in scope, both budget and time. We do not perform neural architecture search or extensive hyperparameter tuning, and we restrict ourselves to a single model size (23M parameters). Scaling up model size, exploring alternative architecture choices, and performing systematic ablations on hyperparameters would likely yield substantial further gains.

Finally, SIFI is applied only at inference time. Incorporating similar coarse-to-dense interpolation strategies directly into the architecture or training objective could allow the model to exploit long contexts more effectively in a single forward pass.

Our current implementation produces point forecasts trained with a Huber loss. This is well aligned with conditional mean forecasting and yields strong performance under MSE, but is less optimal under other losses. Extending Tiny-TSM with a distributional forecast head (e.g., a mixture-of-$t$ head trained with a negative log-likelihood loss, or a quantile head trained with a pinball-Huber loss) would enable flexible trade-offs between metrics by extracting different statistics (mean, median, quantiles) from the predicted distribution. We hypothesize that such distributional training could also improve point forecasts by providing richer gradient signals.

\newpage

\bibliographystyle{plainnat}
\bibliography{main}

@article{Che2016RecurrentNN,
  title={Recurrent Neural Networks for Multivariate Time Series with Missing Values},
  author={Zhengping Che and S. Purushotham and Kyunghyun Cho and David A. Sontag and Yan Liu},
  journal={Scientific Reports},
  year={2016},
  volume={8},
  url={https://api.semanticscholar.org/CorpusID:4900015}
}

@article{StockWatson2002,
author = {Stock, James and Watson, Mark},
year = {2002},
month = {02},
pages = {147-62},
title = {Macroeconomic Forecasting Using Diffusion Indexes},
volume = {20},
journal = {Journal of Business \& Economic Statistics},
doi = {10.1198/073500102317351921}
}

@article{dimri2020time,
  title={Time series analysis of climate variables using seasonal ARIMA approach},
  author={Dimri, Tripti and Ahmad, Shamshad and Sharif, Mohammad},
  journal={Journal of Earth System Science},
  volume={129},
  number={1},
  pages={149},
  year={2020},
  publisher={Springer}
}

@article{gasparin2022deep,
  title={Deep learning for time series forecasting: The electric load case},
  author={Gasparin, Alberto and Lukovic, Slobodan and Alippi, Cesare},
  journal={CAAI Transactions on Intelligence Technology},
  volume={7},
  number={1},
  pages={1--25},
  year={2022},
  publisher={Wiley Online Library}
}

@article{nabavi2024deep,
  title={Deep learning modeling in electricity load forecasting: Improved accuracy by combining DWT and LSTM},
  author={Nabavi, Seyed Azad and Mohammadi, Sahar and Motlagh, Naser Hossein and Tarkoma, Sasu and Geyer, Philipp},
  journal={Energy Reports},
  volume={12},
  pages={2873--2900},
  year={2024},
  publisher={Elsevier}
}

@article{liu2020traffic,
  title={Traffic flow combination forecasting method based on improved LSTM and ARIMA},
  author={Liu, Boyi and Tang, Xiangyan and Cheng, Jieren and Shi, Pengchao},
  journal={International Journal of Embedded Systems},
  volume={12},
  number={1},
  pages={22--30},
  year={2020},
  publisher={Inderscience Publishers (IEL)}
}

@article{liu2020deep,
  title={Deep anomaly detection for time-series data in industrial IoT: A communication-efficient on-device federated learning approach},
  author={Liu, Yi and Garg, Sahil and Nie, Jiangtian and Zhang, Yang and Xiong, Zehui and Kang, Jiawen and Hossain, M Shamim},
  journal={IEEE Internet of Things Journal},
  volume={8},
  number={8},
  pages={6348--6358},
  year={2020},
  publisher={IEEE}
}

@article{belay2023unsupervised,
  title={Unsupervised anomaly detection for IoT-based multivariate time series: Existing solutions, performance analysis and future directions},
  author={Belay, Mohammed Ayalew and Blakseth, Sindre Stenen and Rasheed, Adil and Salvo Rossi, Pierluigi},
  journal={Sensors},
  volume={23},
  number={5},
  pages={2844},
  year={2023},
  publisher={MDPI}
}

@article{Spyros2021,
title = "The M5 competition: Background, organization, and implementation",
author = "Spyros Makridakis and Evangelos Spiliotis and Vassilios Assimakopoulos",
note = "Publisher Copyright: {\textcopyright} 2021 The Authors",
year = "2021",
doi = "10.1016/j.ijforecast.2021.07.007",
language = "English",
journal = "International Journal of Forecasting",
issn = "0169-2070",
publisher = "Elsevier B.V.",
}

@article{sartipi2025benchmarking,
  title={Benchmarking Pre-Trained Time Series Models for Electricity Price Forecasting},
  author={Sartipi, Timoth{\'e}e Hornek Amir and Tchappi, Igor and Fridgen, Gilbert},
  journal={arXiv preprint arXiv:2506.08113},
  year={2025}
}

@article{kugler2015featureless,
  title={Featureless classification of light curves},
  author={K{\"u}gler, Sven Dennis and Gianniotis, Nikos and Polsterer, Kai Lars},
  journal={Monthly Notices of the Royal Astronomical Society},
  volume={451},
  number={4},
  pages={3385--3392},
  year={2015},
  publisher={Oxford University Press}
}

@article{cabrera2024atat,
  title={ATAT: Astronomical Transformer for time series and Tabular data},
  author={Cabrera-Vives, G and Moreno-Cartagena, D and Astorga, N and Reyes-Jainaga, I and F{\"o}rster, F and Huijse, P and Arredondo, J and Arancibia, AM Mu{\~n}oz and Bayo, A and Catelan, M and others},
  journal={Astronomy \& Astrophysics},
  volume={689},
  pages={A289},
  year={2024},
  publisher={EDP Sciences}
}

@misc{aksu2024giftevalbenchmarkgeneraltime,
      title={GIFT-Eval: A Benchmark For General Time Series Forecasting Model Evaluation}, 
      author={Taha Aksu and Gerald Woo and Juncheng Liu and Xu Liu and Chenghao Liu and Silvio Savarese and Caiming Xiong and Doyen Sahoo},
      year={2024},
      eprint={2410.10393},
      archivePrefix={arXiv},
      primaryClass={cs.LG},
      url={https://arxiv.org/abs/2410.10393}, 
}

@misc{auer2025tirexzeroshotforecastinglong,
      title={TiRex: Zero-Shot Forecasting Across Long and Short Horizons with Enhanced In-Context Learning}, 
      author={Andreas Auer and Patrick Podest and Daniel Klotz and Sebastian Böck and Günter Klambauer and Sepp Hochreiter},
      year={2025},
      eprint={2505.23719},
      archivePrefix={arXiv},
      primaryClass={cs.LG},
      url={https://arxiv.org/abs/2505.23719}, 
}

@misc{das2024decoderonlyfoundationmodeltimeseries,
      title={A decoder-only foundation model for time-series forecasting}, 
      author={Abhimanyu Das and Weihao Kong and Rajat Sen and Yichen Zhou},
      year={2024},
      eprint={2310.10688},
      archivePrefix={arXiv},
      primaryClass={cs.CL},
      url={https://arxiv.org/abs/2310.10688}, 
}

@misc{woo2024unifiedtraininguniversaltime,
      title={Unified Training of Universal Time Series Forecasting Transformers}, 
      author={Gerald Woo and Chenghao Liu and Akshat Kumar and Caiming Xiong and Silvio Savarese and Doyen Sahoo},
      year={2024},
      eprint={2402.02592},
      archivePrefix={arXiv},
      primaryClass={cs.LG},
      url={https://arxiv.org/abs/2402.02592}, 
}

@misc{grinsztajn2025tabpfn25advancingstateart,
      title={TabPFN-2.5: Advancing the State of the Art in Tabular Foundation Models}, 
      author={Léo Grinsztajn and Klemens Flöge and Oscar Key and Felix Birkel and Philipp Jund and Brendan Roof and Benjamin Jäger and Dominik Safaric and Simone Alessi and Adrian Hayler and Mihir Manium and Rosen Yu and Felix Jablonski and Shi Bin Hoo and Anurag Garg and Jake Robertson and Magnus Bühler and Vladyslav Moroshan and Lennart Purucker and Clara Cornu and Lilly Charlotte Wehrhahn and Alessandro Bonetto and Bernhard Schölkopf and Sauraj Gambhir and Noah Hollmann and Frank Hutter},
      year={2025},
      eprint={2511.08667},
      archivePrefix={arXiv},
      primaryClass={cs.LG},
      url={https://arxiv.org/abs/2511.08667}, 
}

@misc{hollmann2023tabpfntransformersolvessmall,
      title={TabPFN: A Transformer That Solves Small Tabular Classification Problems in a Second}, 
      author={Noah Hollmann and Samuel Müller and Katharina Eggensperger and Frank Hutter},
      year={2023},
      eprint={2207.01848},
      archivePrefix={arXiv},
      primaryClass={cs.LG},
      url={https://arxiv.org/abs/2207.01848}, 
}

@misc{oreshkin2020nbeatsneuralbasisexpansion,
      title={N-BEATS: Neural basis expansion analysis for interpretable time series forecasting}, 
      author={Boris N. Oreshkin and Dmitri Carpov and Nicolas Chapados and Yoshua Bengio},
      year={2020},
      eprint={1905.10437},
      archivePrefix={arXiv},
      primaryClass={cs.LG},
      url={https://arxiv.org/abs/1905.10437}, 
}

@misc{ansari2025chronos2univariateuniversalforecasting,
      title={Chronos-2: From Univariate to Universal Forecasting}, 
      author={Abdul Fatir Ansari and Oleksandr Shchur and Jaris Küken and Andreas Auer and Boran Han and Pedro Mercado and Syama Sundar Rangapuram and Huibin Shen and Lorenzo Stella and Xiyuan Zhang and Mononito Goswami and Shubham Kapoor and Danielle C. Maddix and Pablo Guerron and Tony Hu and Junming Yin and Nick Erickson and Prateek Mutalik Desai and Hao Wang and Huzefa Rangwala and George Karypis and Yuyang Wang and Michael Bohlke-Schneider},
      year={2025},
      eprint={2510.15821},
      archivePrefix={arXiv},
      primaryClass={cs.LG},
      url={https://arxiv.org/abs/2510.15821}, 
}

@misc{hoo2025tablestimetabpfnv2outperforms,
      title={From Tables to Time: How TabPFN-v2 Outperforms Specialized Time Series Forecasting Models}, 
      author={Shi Bin Hoo and Samuel Müller and David Salinas and Frank Hutter},
      year={2025},
      eprint={2501.02945},
      archivePrefix={arXiv},
      primaryClass={cs.LG},
      url={https://arxiv.org/abs/2501.02945}, 
}

@misc{nie2023timeseriesworth64,
      title={A Time Series is Worth 64 Words: Long-term Forecasting with Transformers}, 
      author={Yuqi Nie and Nam H. Nguyen and Phanwadee Sinthong and Jayant Kalagnanam},
      year={2023},
      eprint={2211.14730},
      archivePrefix={arXiv},
      primaryClass={cs.LG},
      url={https://arxiv.org/abs/2211.14730}, 
}

@misc{ansari2024chronoslearninglanguagetime,
      title={Chronos: Learning the Language of Time Series}, 
      author={Abdul Fatir Ansari and Lorenzo Stella and Caner Turkmen and Xiyuan Zhang and Pedro Mercado and Huibin Shen and Oleksandr Shchur and Syama Sundar Rangapuram and Sebastian Pineda Arango and Shubham Kapoor and Jasper Zschiegner and Danielle C. Maddix and Hao Wang and Michael W. Mahoney and Kari Torkkola and Andrew Gordon Wilson and Michael Bohlke-Schneider and Yuyang Wang},
      year={2024},
      eprint={2403.07815},
      archivePrefix={arXiv},
      primaryClass={cs.LG},
      url={https://arxiv.org/abs/2403.07815}, 
}

@misc{cohen2024tototimeseriesoptimized,
      title={Toto: Time Series Optimized Transformer for Observability}, 
      author={Ben Cohen and Emaad Khwaja and Kan Wang and Charles Masson and Elise Ramé and Youssef Doubli and Othmane Abou-Amal},
      year={2024},
      eprint={2407.07874},
      archivePrefix={arXiv},
      primaryClass={cs.LG},
      url={https://arxiv.org/abs/2407.07874}, 
}

@misc{xie2025caukerclassificationtimeseries,
      title={CauKer: classification time series foundation models can be pretrained on synthetic data only}, 
      author={Shifeng Xie and Vasilii Feofanov and Marius Alonso and Ambroise Odonnat and Jianfeng Zhang and Themis Palpanas and Ievgen Redko},
      year={2025},
      eprint={2508.02879},
      archivePrefix={arXiv},
      primaryClass={cs.LG},
      url={https://arxiv.org/abs/2508.02879}, 
}

@misc{bommasani2022opportunitiesrisksfoundationmodels,
      title={On the Opportunities and Risks of Foundation Models}, 
      author={Rishi Bommasani and Drew A. Hudson and Ehsan Adeli and Russ Altman and Simran Arora and Sydney von Arx and Michael S. Bernstein and Jeannette Bohg and Antoine Bosselut and Emma Brunskill and Erik Brynjolfsson and Shyamal Buch and Dallas Card and Rodrigo Castellon and Niladri Chatterji and Annie Chen and Kathleen Creel and Jared Quincy Davis and Dora Demszky and Chris Donahue and Moussa Doumbouya and Esin Durmus and Stefano Ermon and John Etchemendy and Kawin Ethayarajh and Li Fei-Fei and Chelsea Finn and Trevor Gale and Lauren Gillespie and Karan Goel and Noah Goodman and Shelby Grossman and Neel Guha and Tatsunori Hashimoto and Peter Henderson and John Hewitt and Daniel E. Ho and Jenny Hong and Kyle Hsu and Jing Huang and Thomas Icard and Saahil Jain and Dan Jurafsky and Pratyusha Kalluri and Siddharth Karamcheti and Geoff Keeling and Fereshte Khani and Omar Khattab and Pang Wei Koh and Mark Krass and Ranjay Krishna and Rohith Kuditipudi and Ananya Kumar and Faisal Ladhak and Mina Lee and Tony Lee and Jure Leskovec and Isabelle Levent and Xiang Lisa Li and Xuechen Li and Tengyu Ma and Ali Malik and Christopher D. Manning and Suvir Mirchandani and Eric Mitchell and Zanele Munyikwa and Suraj Nair and Avanika Narayan and Deepak Narayanan and Ben Newman and Allen Nie and Juan Carlos Niebles and Hamed Nilforoshan and Julian Nyarko and Giray Ogut and Laurel Orr and Isabel Papadimitriou and Joon Sung Park and Chris Piech and Eva Portelance and Christopher Potts and Aditi Raghunathan and Rob Reich and Hongyu Ren and Frieda Rong and Yusuf Roohani and Camilo Ruiz and Jack Ryan and Christopher Ré and Dorsa Sadigh and Shiori Sagawa and Keshav Santhanam and Andy Shih and Krishnan Srinivasan and Alex Tamkin and Rohan Taori and Armin W. Thomas and Florian Tramèr and Rose E. Wang and William Wang and Bohan Wu and Jiajun Wu and Yuhuai Wu and Sang Michael Xie and Michihiro Yasunaga and Jiaxuan You and Matei Zaharia and Michael Zhang and Tianyi Zhang and Xikun Zhang and Yuhui Zhang and Lucia Zheng and Kaitlyn Zhou and Percy Liang},
      year={2022},
      eprint={2108.07258},
      archivePrefix={arXiv},
      primaryClass={cs.LG},
      url={https://arxiv.org/abs/2108.07258}, 
}

@misc{brown2020languagemodelsfewshotlearners,
      title={Language Models are Few-Shot Learners}, 
      author={Tom B. Brown and Benjamin Mann and Nick Ryder and Melanie Subbiah and Jared Kaplan and Prafulla Dhariwal and Arvind Neelakantan and Pranav Shyam and Girish Sastry and Amanda Askell and Sandhini Agarwal and Ariel Herbert-Voss and Gretchen Krueger and Tom Henighan and Rewon Child and Aditya Ramesh and Daniel M. Ziegler and Jeffrey Wu and Clemens Winter and Christopher Hesse and Mark Chen and Eric Sigler and Mateusz Litwin and Scott Gray and Benjamin Chess and Jack Clark and Christopher Berner and Sam McCandlish and Alec Radford and Ilya Sutskever and Dario Amodei},
      year={2020},
      eprint={2005.14165},
      archivePrefix={arXiv},
      primaryClass={cs.CL},
      url={https://arxiv.org/abs/2005.14165}, 
}

@misc{radford2021learningtransferablevisualmodels,
      title={Learning Transferable Visual Models From Natural Language Supervision}, 
      author={Alec Radford and Jong Wook Kim and Chris Hallacy and Aditya Ramesh and Gabriel Goh and Sandhini Agarwal and Girish Sastry and Amanda Askell and Pamela Mishkin and Jack Clark and Gretchen Krueger and Ilya Sutskever},
      year={2021},
      eprint={2103.00020},
      archivePrefix={arXiv},
      primaryClass={cs.CV},
      url={https://arxiv.org/abs/2103.00020}, 
}

@misc{he2021maskedautoencodersscalablevision,
      title={Masked Autoencoders Are Scalable Vision Learners}, 
      author={Kaiming He and Xinlei Chen and Saining Xie and Yanghao Li and Piotr Dollár and Ross Girshick},
      year={2021},
      eprint={2111.06377},
      archivePrefix={arXiv},
      primaryClass={cs.CV},
      url={https://arxiv.org/abs/2111.06377}, 
}

@misc{liu2025moirai20timeseries,
      title={Moirai 2.0: When Less Is More for Time Series Forecasting}, 
      author={Chenghao Liu and Taha Aksu and Juncheng Liu and Xu Liu and Hanshu Yan and Quang Pham and Doyen Sahoo and Caiming Xiong and Silvio Savarese and Junnan Li},
      year={2025},
      eprint={2511.11698},
      archivePrefix={arXiv},
      primaryClass={cs.LG},
      url={https://arxiv.org/abs/2511.11698}, 
}


\appendix

\section{Appendix / supplemental material}

\subsection{Base Time series Generators}

We group base generators by qualitative pattern and provide visual examples in Figure~\ref{fig:BaseGen}. Each group may contain multiple implementations (e.g., several types of noise).

\begin{itemize}
    \item Sum-of-sinusoids

    \item Linear and non-linear trends

    \item Periodic random walks

    \item Trends with random resets to zero

    \item Modular circular re-indexing of existing base generators: given a series $(x_t)_{t=1, ..., L}$ of length $L$ and an integer $k$, we construct a new index sequence
    \begin{equation}
        t'_i = (k \cdot i) \bmod L, \qquad i = 1,2,\dots,
    \end{equation}
    and define the derived series $(x_{t'_i})_i$.

    \item Periodic processes with intermittent explosive episodes

    \item Periodic time series with phase shifts

    \item Periodic non-negative time series with a floor at zero

    \item Integer-count time series

    \item Approximately periodic integer time series

    \item Mixtures of other periodic time series from periodic Base Generators

    \item Noise processes (Gaussian noise, random walks, uniform noise, etc.)

    \item Noisy auto-regressive processes

    \item Explosive processes: Stochastic processes possessing periodicity and localized episodes of explosive scaling

    \item Local trend processes

\begin{figure}
    \centering
    \includegraphics[width=0.8\linewidth]{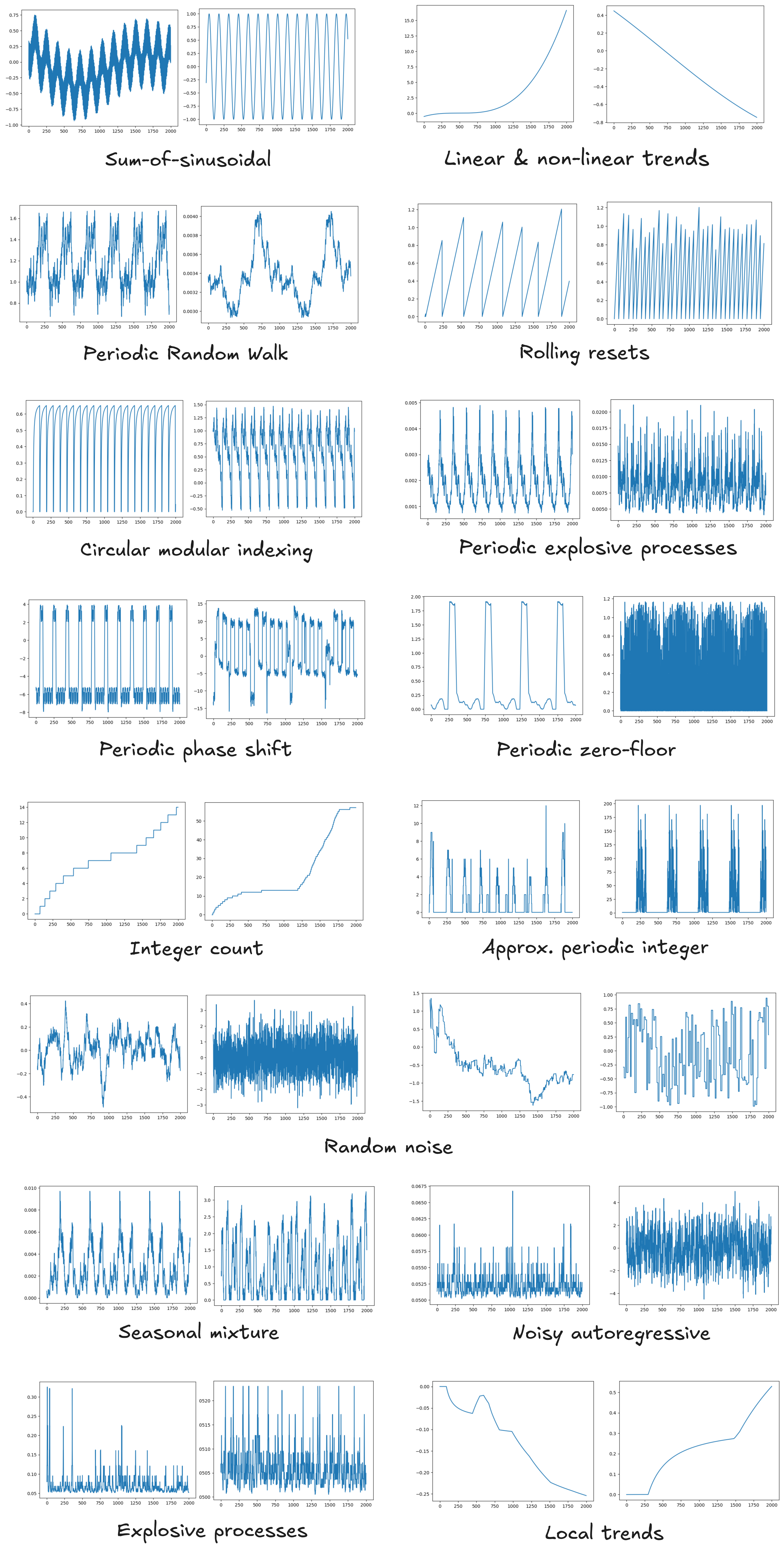}
    \caption{Selected examples for time series generated by different groups of Base Generators. We include 2 examples each, except for noise, which has 4 examples due to its substantial importance.}
    \label{fig:BaseGen}
\end{figure}

\begin{figure}
    \centering
    \includegraphics[width=0.825\linewidth]{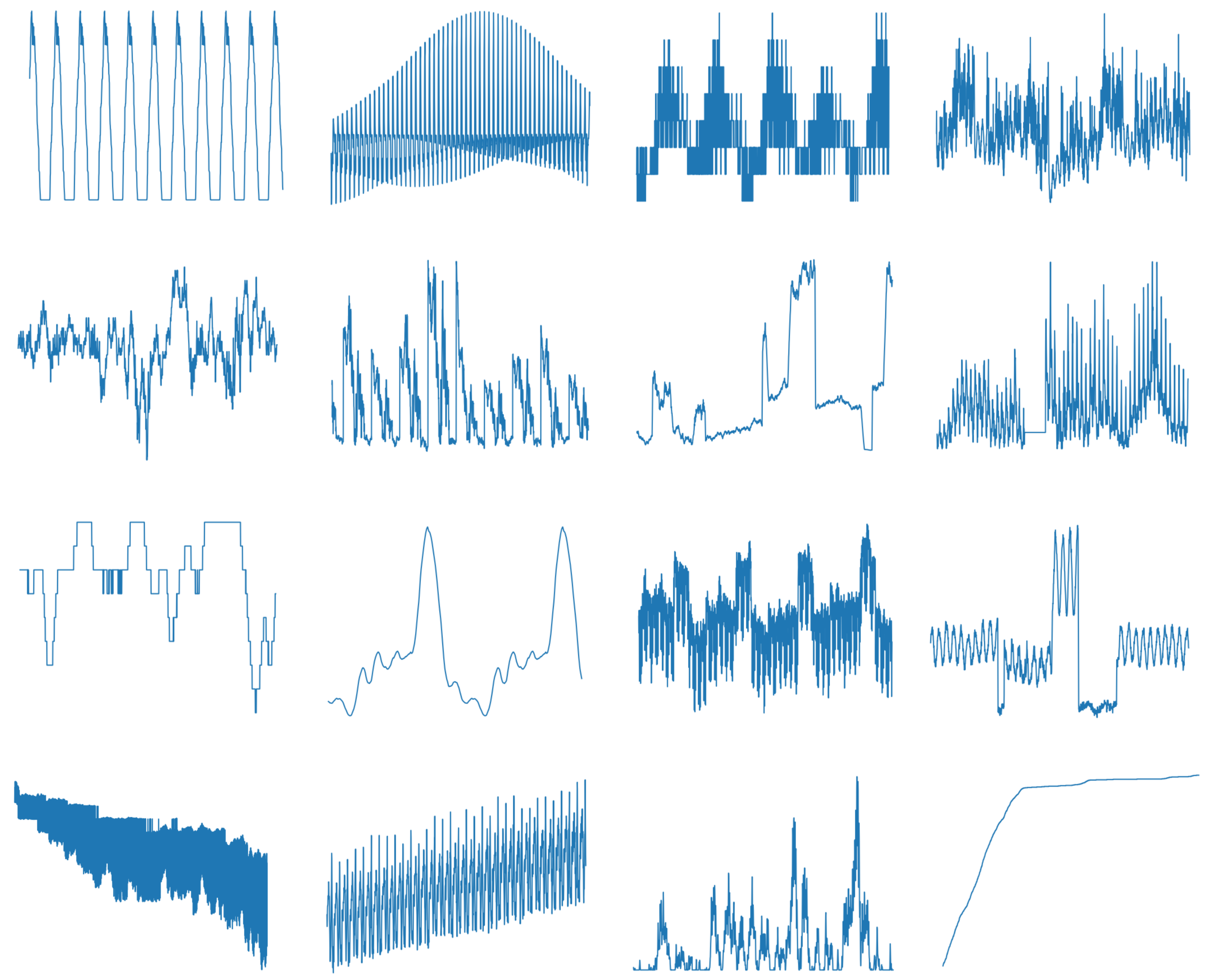}
    \caption{Selected examples from SynthTS. Unlike the samples from the base generators visualized above, these examples are outputs of the full SynthTS pipeline.}
    \label{fig:TSSamples}
\end{figure}

\end{itemize}


\end{document}